\documentclass[11pt,a4paper]{article}
\usepackage[hyperref]{naaclhlt2018}
\usepackage{times}
\usepackage{latexsym}
\usepackage{graphicx}
\usepackage{amssymb}
\usepackage{amsmath}

\usepackage{url}

\aclfinalcopy 

\title{Co-Attention Based Neural Network for Source-Dependent Essay Scoring}

\author{Haoran Zhang \\
  Department of Computer Science\\
  University of Pittsburgh \\
  Pittsburgh, PA 15260 \\
  {\tt colinzhang@cs.pitt.edu} \\\And
  Diane Litman \\
  Department of Computer Science \& LRDC\\
 University of Pittsburgh \\
  Pittsburgh, PA 15260 \\
  {\tt litman@cs.pitt.edu} \\}

\date{}

\begin{document}
\maketitle
\begin{abstract}
This paper presents an investigation of using a co-attention based neural network for source-dependent essay scoring. We use a co-attention mechanism to help the model learn the importance of each part of the essay more accurately. Also, this paper shows that the co-attention based neural network model provides reliable score prediction of source-dependent responses. We evaluate our model on two source-dependent response corpora. Results show that our model outperforms the baseline on both corpora. We also show that the attention of the model is similar to the expert opinions with examples. 
\end{abstract}

\section{Introduction}

Manually grading students' essays is labor intensive. Therefore, many automated essay scoring (AES) methods have been developed to support grading essays at scale. However, in different grading tasks, the information required by an AES system is different. For example, if a system needs to assign a holistic score to the essay, the system needs to take all information into account. In contrast, if a system needs to assign a score for one specific aspect of the essay (e.g. use of evidence), the system needs to ignore some information. Also, if an essay is a source-dependent essay, the system needs to exploit knowledge of the source article.


This paper focuses on  source-dependent essay assessment. In this task, students read a source article before writing the essay, and assessment involves recognizing and analyzing references to the article in the essay. We propose a new type of co-attention based neural network model tailored to source-dependent grading, then use two source-dependent essay corpora to evaluate our model. Our first corpus contains the four source-dependent essay sets in the Automated Student Assessment Prize (ASAP) corpus\footnote{https://www.kaggle.com/c/asap-aes}.
The ASAP grading task is to assign a holistic score to each essay. The second corpus uses the Response to Text Assessment (RTA)  \cite{correnti2013assessing} to assess students' analytic writing skills. Instead of evaluating holistic writing skills, the RTA was designed to evaluate students' writing skills along five dimensions: Analysis, Evidence, Organization, Style, and MUGS (Mechanics, Usage, Grammar, and Spelling). Our grading task for this corpus is to assign an Evidence score to each essay, by evaluating students' ability to find and use evidence from a source article to support their claims.



The main contributions of this paper are as follows. First, we introduce a co-attention based neural network model that is fully automated and does not need any expert effort to encode knowledge of a source article. Second, our co-attention based neural network model extends prior work by designing the model to take a source article into account during grading. Third, we apply our model to the subset of source-dependent responses tasks in the ASAP corpus and show that the model outperforms a previous neural network model developed for the full corpus. Fourth, we show that our model also performs well on the RTA task and again significantly outperforms our baseline neural net model. Last, we use examples to show that our model can assign reasonable attention scores to different sentences in the essay.

In the following sections, we first present related research. Then we describe our tasks by introducing the ASAP corpus and the RTA corpus. Next, we explain the structure of our co-attention based neural network model. Finally, we discuss the results of our experiments and future plans.

\section{Related Work}

Previous research in AES needed feature engineering. In very early work, \citet{page1968use} developed an AES tool named Project Essay Grade (PEG) by only using linguistic surface features. A more recent well-known AES system is E-Rater \cite{burstein1998automated}, which employs many more natural language processing (NLP) technologies.
Later, \citet{attali2004automated} released E-Rater V2, where they created a new set of features  to represent linguistic characteristic related to organization and development, lexical complexity, prompt-specific vocabulary usage, etc. Similarly to \citet{page1968use}, this system  used regression equations for assessment of student essays. 
One limitation of all of the above models is that  all  need handcrafted features for training the model. 
In contrast, our model uses a neural network for the AES task and thus does not require feature engineering. 

Recently,  neural network models have been introduced into AES, making the development of handcrafted features unnecessary or at least optional.
\citet{alikaniotis2016automatic} and \citet{taghipour2016neural} presented AES models that used Long Short Term Memory (LSTM) networks. Differently, \citet{dong2016automatic} used a Convolutional Neural Network (CNN) model for essay scoring by applying two CNN layers on both the word level and then sentence level. Later, \citet{dong2017attention} presented another work that uses attention pooling to replace the mean over time pooling after the convolutional layer in both word level and sentence levels. However, none of these neural network grading models consider the source article if it exists. In this paper, we introduce a neural network model that takes the source article into account by using a co-attention mechanism instead of the self-attention mechanism of prior work.

Our work not only focuses on essay assessment using a holistic score, but also evaluates a particular dimension of argument-oriented writing skills, namely use of Evidence. 
\citet{louis2010off} analyze only the content of essays by detecting off-topic essays. \citet{ong2014ontology} used argumentation mining techniques to evaluate if students use enough evidence to support their positions. However, these two prior studies are not suitable for our task because they did not measure the use of content or evidence from a source article. With respect to source-based dimensional essay analysis, \citet{rahimi2014automatic, rahimi2017assessing} developed a set of rubric-based features that compared a student's essay and a source article in terms of number of related words or paraphrases. \citet{zhang2017word} improved their  model by introducing word embedding into the feature extraction process to extract relationships previously missed due to lexical errors or use of different vocabulary. However, in both of these studies, human effort was still necessary for pre-processing the source article, for example, by having experts manually create a list of important words and phrases in the article which  the system would compare with features extracted from the student's essay. In contrast, our work does not need any human effort to analyze the source article before essay grading. Although \citet{rahimi2016automatically} investigated extracting example lists by using LDA \citep{blei2003latent} model, the data-driven model missed an example when there was no essay mentioning the example. \citet{klebanov2014content} predicted which parts of the source material were important and that students needed to use in their essays. The essay score is required to obtain the content importance for their work, but our work does not need to know the essay score while identifying the content importance.

\section{Data}

We use two different essay corpora in our experiments: source-based essays from the ASAP corpus, and source-based RTA essays. While the full ASAP corpus contains essays in response to 8 different prompts, we  use only essays in response to the 4 source-dependent prompts. The gold standard ASAP assessment is a holistic score. In contrast, the gold standard assessment in the RTA corpus is an Evidence score. In particular, the assessment only considers how students use evidence from a source article to support their claims; the assessement thus ignores the lexical and syntactic mistakes made by students and the organization of the essay when assessing the evidence dimension.

\subsection{ASAP}

\begin{figure}[h]
\begin{quote}
{\bf Source  Excerpt: }My mother and father had come to this country with such courage, without any knowledge of the language or the culture. They came selflessly, as many immigrants do, to give their children a better life, even though it meant leaving behind their families, friends, and careers in the country they loved.

{\bf Essay Prompt: } Describe the mood created by the author in the memoir. Support your answer with relevant and specific information from the memoir.
\end{quote}
\caption{A source excerpt for ASAP Prompt 5.}
\label{fig:asapprompt}
\end{figure}

The Automated Student Assessment Prize (ASAP) corpus consists of written responses to 8 prompts. Among them, prompts 3, 4, 5, and 6 are source-dependent  which means students read an article before writing their essays. Since the scores assigned to essays are holistic,  assessment considers the overall quality of the essay, not just a specific dimension. Figure~\ref{fig:asapprompt} contains an excerpt from an ASAP source article and the associated Prompt 5.

\begin{table}[h]
\begin{center}
\begin{tabular}{|r|cccc|}
\hline \bf Prompt & \bf 3 & \bf4 & \bf 5 & \bf 6 \\ \hline
\bf Score 0 & 39 & 311 & 24 & 44 \\
& (2\%) & (18\%) & (1\%) & (3\%) \\
\bf Score 1 & 607 & 636 & 302 & 167 \\
& (35\%) & (36\%) & (17\%) & (9\%) \\
\bf Score 2 & 657 & 570 & 649 & 405 \\
& (38\%) & (32\%) & (36\%) & (23\%) \\
\bf Score 3 & 423 & 253 & 572 & 817 \\
& (25\%) & (14\%) & (32\%) & (45\%) \\
\bf Score 4 & NA & NA & 258 & 367 \\
& & & (14\%) & (20\%) \\ \hline
\bf Total & 1726 & 1770 & 1805 & 1800 \\
\hline
\end{tabular}
\end{center}
\caption{\label{distribution2} The holistic score distribution of ASAP.}
\end{table}

In this paper, we only focus on prompts 3, 4, 5, and 6 (denoted by $ASAP_3$, $ASAP_4$, $ASAP_5$, and $ASAP_6$ respectively), because they are source-dependent responses. In ASAP, different prompts have different score ranges. The score range of $ASAP_3$ and $ASAP_4$ is 0 to 3, while  the range of $ASAP_5$ and $ASAP_6$ is 0 to 4. Figure~\ref{fig:exampleessayasap} shows an excerpt of an essay with score of 4 for $ASAP_5$. The score distribution is shown in Table~\ref{distribution2}. 

\begin{figure}[h]
\begin{quote}

{\bf Essay Excerpt: }The author of the memoir, Narciso Rodriguez creates a caring, happy, and thoughtful mood. By mentioning the Cuban traditions shared in the neighborhood between close friends, and cooking in the kitchen to share a great meal with one another the mood is happy. When Narciso talks about the great friends he made from different heritages and knowing the entire community like family the mood is thoughtful and caring because it shows that the people really appreciated each other's company...
\end{quote}
\caption{Excerpt of an essay with score of 4 for ASAP Prompt 5.}
\label{fig:exampleessayasap}
\end{figure}

\subsection{RTA}

The RTA corpora were collected from upper elementary level students, as described by \citet{correnti2013assessing}. There are two forms of RTA based on different articles that students read before writing essays. The first article is from \emph{Time for Kids} about the Millennium Villages Project, an effort by the United Nations to end poverty in a rural village in Sauri, Kenya; we refer to it as $RTA_{MVP}$. The other article talks about the importance of space exploration; we refer to refer it as $RTA_{Space}$. Figure~\ref{fig:rtaprompt} shows an excerpt from the $RTA_{MVP}$ article and the associated essay writing prompt. Bolded text spans in the article excerpt are pieces of evidence that our experts (School of Education RTA team members) manually labeled as being important for students to include in their essays. 

\begin{figure}[h]
\begin{quote}
{\bf Source Excerpt: }Today, Yala Sub-District {\bf Hospital has medicine}, {\bf free of charge}, {\bf for all of the most common diseases}. {\bf Water is connected to the hospital}, which also has a {\bf generator for electricity}. {\bf Bed nets are used} in every sleeping site in Sauri...

{\bf Essay Prompt: } The author provided one specific example of how the quality of life can be improved by the Millennium Villages Project in Sauri, Kenya. Based on the article, did the author provide a convincing argument that winning the fight against poverty is achievable in our lifetime? Explain why or why not with 3-4 examples from the text to support your answer.
\end{quote}
\caption{A source excerpt for the $RTA_{MVP}$ prompt.}
\label{fig:rtaprompt}
\end{figure}


Evidence usage in each RTA essay was scored on a scale of 1 to 4 (low to high). The distribution of Evidence scores is shown in Table~\ref{distribution}. Figure~\ref{fig:exampleessay} shows a student essay with a score of 3. Our experts manually bolded all pieces of evidence found in this essay.

\begin{figure}[h]
\begin{quote}

{\bf Essay: }In my opinion I think that they will {\bf achieve it in lifetime}. During the years threw {\bf 2004 and 2008 they made progress}. People didn’t have the money to buy the stuff in 2004. {\bf The hospital was packed with patients} and they didn’t have alot of treatment in 2004. In 2008 it changed the {\bf hospital had medicine}, {\bf free of charge}, and {\bf for all the common dieases}. {\bf Water was connected to the hospital} and has a {\bf generator for electricity}. {\bf Everybody has net} in their site. {\bf The hunger crisis has been addressed} with {\bf fertilizer and seeds}, as well as the {\bf tools needed to maintain the food}. {\bf The school has no fees} and {\bf they serve lunch}. To me that’s sounds like it is going achieve it in the lifetime.
\end{quote}
\caption{A $RTA_{MVP}$ essay with score of 3.}
\label{fig:exampleessay}
\end{figure}

\begin{table}[h]
\begin{center}
\begin{tabular}{|r|cc|}
\hline \bf Prompt & \bf $RTA_{MVP}$ & \bf $RTA_{Space}$  \\ \hline
\bf  Score 1 & 852 & 538 \\
& (29\%) & (26\%) \\
\bf Score 2 & 1197 & 789 \\
& (40\%) & (38\%) \\
\bf Score 3 & 616 & 512 \\
& (21\%) & (25\%) \\
\bf Score 4 & 305 & 237 \\
& (10\%) & (11\%) \\ \hline
\bf Total & 2970 & 2076 \\
\hline
\end{tabular}
\end{center}
\caption{\label{distribution} The Evidence score distribution of RTA.}
\end{table}

\begin{figure*}[h]
\includegraphics[width=\textwidth, height=8cm]{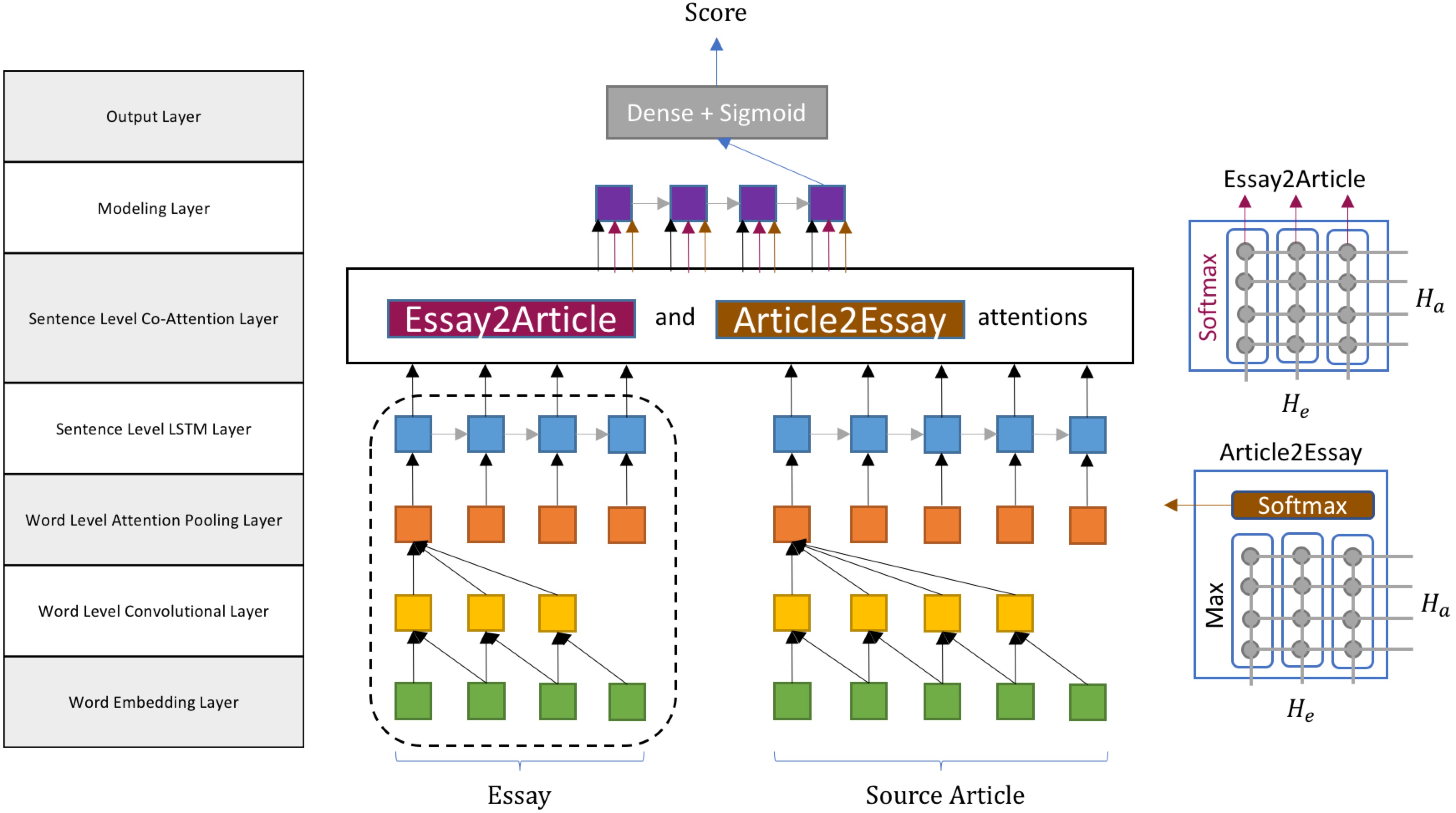}
\caption{The Co-Attention Based Neural Network Structure.}
\label{fig:network}
\end{figure*}

\section{Model}

Our network is inspired by the hierarchical neural network model presented by \citet{dong2017attention}. In their model, they considered each essay as a sequence of sentences rather than a sequence of words. Their model has three parts. First, they used a convolutional layer and attention pooling layer to get sentence representation. Second, they used an LSTM layer and another attention pooling layer for document representation. Finally, they used a sigmoid layer for score prediction.

Differently from their model, our model replaces the attention pooling layer for document representation with a bi-directional attention flow layer and an additional modeling layer \citep{seo2016bidirectional}. By doing so, our model considers students' essays associated with a source article and this attention mechanism captures the relationship between the essay and the source article. In particular, a higher attention score will be assigned to  sentences that are mentioned in the article but less mentioned in other essays. Our model is a hierarchical neural network and consists of seven layers. Figure~\ref{fig:network} shows the structure of our network. The layers in the dashed box were presented by \citet{dong2017attention}. The sentence level co-attention layer was presented by \citet{seo2016bidirectional}. 

\subsection{Word Embedding Layer}

This layer maps each word in  sentences to a high dimension vector. We use the GloVe pre-trained word embeddings \citep{pennington2014glove} to obtain the word embedding vector for each word. It was trained on 6 billion words from Wikipedia 2014 and Gigaword 5. It has 400,000 uncased vocabulary items. The dimensionality of GloVe in our model is 50 dimensions. The outputs of this layer are two matrices, $L_E \in \mathbb{R}^{S_e \times W_e \times d_L}$ for the essay and $L_A \in \mathbb{R}^{S_a \times W_a \times d_L}$ for the article, where $S_e$, $S_a$, $W_e$, $W_a$, and $d_L$ are number of sentences of the essay and the article, length of sentences of the essay and the article, and the embedding size, respectively. Same to \citet{dong2017attention}, a dropout is applied after the word embedding layer.

\subsection{Word Level Convolutional Layer}
In this layer, we perform 1D convolution over the word representations of both $L_E$ and $L_A$, so that we can get local representation of each sentence. For each word $w_i$ in each sentence, we perform 1D convolution:
\[
p_i = g([w_i:w_{i+k-1}] \cdot U_p + b_p) \tag{1}
\]
where $g$ is a nonlinear activation, $k$ is the kernel size, $U_p$ is the filter weight matrix, and $b_p$ is the bias vector. The outputs of this layer are $C_e \in \mathbb{R}^{S_e \times P_e \times d_C}$ for the essay and $C_a \in \mathbb{R}^{S_a \times P_a \times d_C}$ for the article, where $P_e$ and $P_a$ are filtered lengths of sentences of the essay and the article, respectively. $d_C$ is the number of filters of the 1D convolution layer.

\subsection{Word Level Attention Pooling Layer}
After the convolutional layer, a pooling layer is demanded to obtain the sentence representations. In this layer, we follow the same design presented by \citet{dong2017attention}. The attention pooling is defined as equations below:
\[
m_i = tanh(U_m \cdot p_i + b_m) \tag{2}
\]
\[
v_i = \frac{e^{u_v \cdot m_i}}{\sum{e^{u_v \cdot m_j}}} \tag{3}
\]
\[
s = \sum{v_i p_i} \tag{4}
\]
where $U_m$, $u_v$ and $b_m$ are weight matrix, vector, and bias vector, respectively. $m_i$ and $v_i$ are attention vector and attention weight for $p_i$. The outputs of this layer are $A_e \in \mathbb{R}^{S_e \times d_C}$ for the essay and $A_a \in \mathbb{R}^{S_a \times d_C}$ for the article.

\subsection{Sentence Level LSTM Layer}
In this layer, we use a Long Short-Term Memory Network (LSTM) \citep{hochreiter1997long} over the sentence representations of the essay and the article to capture contextual evidence from previous sentences to refine the sentence representation.

The LSTM unit is a special kind of RNN unit which has long-term dependency learning ability. LSTMs use three gates to control information flow to avoid the long-term dependency problem by forgetting or remembering information in each LSTM unit. They are an input gate, a forget gate, and an output gate. The following equations define the LSTM unit:
\[
f_t = \sigma(W_f \cdot [h_{t-1},s_t] + b_f) \tag{5}
\]
\[
i_t = \sigma(W_i \cdot [h_{t-1},s_t] + b_i) \tag{6}
\]
\[
\tilde{c}_t = tanh(W_c \cdot [h_{t-1},s_t] + b_c) \tag{7}
\]
\[
c_t = f_t * c_{t-1} + i_t * \tilde{c}_t \tag{8}
\]
\[
o_t = \sigma(W_o \cdot [h_{t-1},s_t] + b_o) \tag{9}
\]
\[
h_t = o_t * tanh(c_t) \tag{10}
\]
where $s_t$ and $h_t$ are the input sentence and the output state of time $t$, respectively. $W_f$, $W_i$, $W_c$, and $W_o$ are weight matrices. $b_f$, $b_i$, $b_c$, and $b_o$ are bias vectors. $\sigma$ is the sigmoid function, and $*$ is element-wise multiplication. The output of this layer are $H_e \in \mathbb{R}^{S_e \times d_H}$ for the essay and $H_a \in \mathbb{R}^{S_a \times d_H}$ for the article, where $d_H$ is the dimensionality of the output.

\subsection{Sentence Level Co-Attention Layer}

The concept of this layer is presented by \citet{seo2016bidirectional} in the part of attention flow layer. This layer links information from $H_e$ and $H_a$, and generates a collection of article aware features vector of essay sentences. The attention is computed in two directions, from essay to article, and vice versa. Both attention scores are figured from a similarity matrix by the following equation:
\[
Sim = W_{sim}^T \cdot [he_t;  ha_j;  ha_t * ha_j^T] + b_{sim} \tag{11}
\]
where $W_{sim}$ is weight matrix, $he_t$ and $ha_j$ are $t_{th}$ row vector of $H_e$ and $j_{th}$ row vector of $H_a$, $b_{sim}$ is bias vector. $*$ is element-wise multiplication. $[;]$ is vector concatenation. After obtaining the similarity matrix $Sim \in \mathbb{R}^{S_e \times S_a}$, we compute the attention in two directions.

{\bf Essay to Article Attention} measures which sentences in the article are similar to each sentence in students' essays. The following equations define the essay to article attention:
\[
a_{ea} = softmax(Sim) \tag{12}
\]
\[
\tilde{H}_a = a_{ea} H_a \tag{13}
\]
where $a_{ea} \in \mathbb{R}^{S_e \times S_a}$ represents the attention score of each sentence in the article associate with each sentence in the essay, $softmax$ is performed across each row. The output of this $\tilde{H}_a \in \mathbb{R}^{S_e \times d_H}$.

{\bf Article to Essay Attention} measures which sentences in the essay have the closest meaning to one of the sentences in the article. The following equations define the article to essay attention:
\[
a_{ae} = softmax(max_{col}(Sim)) \tag{14}
\]
\[
\tilde{h}_e = a_{ae}^T H_e \tag{15}
\]
where $a_{ae} \in \mathbb{R}^{S_e}$, $max_{col}$ is a maximum function performed across the column, and $\tilde{h}_e \in \mathbb{R}^{d_H}$. Because $max_{col}$ will find out which sentence in the article has the closest meaning to each sentence in the essay, so $\tilde{h}_e$ represents the attention score of the most important sentence in the essay associated with the article. After tiling $S_e$ times, the final output of this layer is $\tilde{H}_e \in \mathbb{R}^{S_e \times d_H}$.

The final output $G$ is a concatenated matrix of $H_e$, $\tilde{H}_e$, and $\tilde{H}_a$ defined by:
\[
G = [H_e; \tilde{H}_a; H_e * \tilde{H}_a; H_e * \tilde{H}_e] \tag{16}
\]
where $*$ is element-wise multiplication, and $[;]$ is concatenation, $H_e$ is the original representation of essay, $\tilde{H}_a$ is the essay to article attention, $H_e * \tilde{H}_a$ is the self-aware representation, and $H_e * \tilde{H}_e$ is article-aware representation. Therefore, the output of this layer is $G \in \mathbb{R}^{S_e \times 4d_H}$, the article-aware representation of each sentence in the essay.

\subsection{Modeling Layer}
$G$ is the representation of each sentence, and we need the representation of the essay. Therefore, we introduce another LSTM layer for modeling the essay and only use the output of the final LSTM unit as the output of this layer $M \in \mathbb{R}^{d_M}$, where $d_M$ is the dimensionality of the output of LSTM units. 

\subsection{Output Layer}
After obtaining the essay representation $M$, a linear layer with sigmoid activation will predict the final output. The following equation defines the output layer:
\[
y=sigmoid(W_oM+b_o) \tag{17}
\]
where $W_o$ is weight vector, and $b_o$ is bias vector. $y$ is the final predicted score of the essay.

\section{Training}
{\bf Loss.} \citet{dong2017attention} used mean squared error (MSE) loss, thus we use the same loss function. MSE evaluates the average of squared error between the predicted score and the gold standard. Thus it is widely used in regression tasks. The following equation defines MSE:
\[
mse(y, y')=\frac{1}{N}\sum_{i=1}^{N}{(y_i-y'_i)^2} \tag{18}
\]
where $y_i$ is the predicted score, $y'_i$ is the gold standard, $N$ it the total number of samples.

{\bf Optimization.} The optimizer we use is RMS\-prop \citep{dauphin2015equilibrated}. The initial learning rate is 0.001, momentum is 0.9, and Dropout rate is 0.5 for preventing overfitting. These setting are the same as used by \citet{dong2017attention}.

\section{Experimental Setup}
We configure experiments to test three hypotheses: 

\begin{itemize}
\item[H1:] the model we proposed (denoted by CO-ATTN) will outperform or at least perform equally well as the baseline (denoted by SELF-ATTN) presented by \citet{dong2017attention} on four ASAP essay corpora in the holistic score prediction task.

\item[H2:] the model we proposed will outperform or at least perform equally well as the baseline on two RTA corpora in the Evidence score prediction task. 

\item[H3:] the model we proposed will outperform or at least perform equally well as the non-neural network baselines on both corpora.

\end{itemize}


We use NLTK \citep{bird2009natural} for text preprocessing. The vocabulary size of the data is limited to 4000, and all scores are scaled to the range [0, 1],  following \citet{taghipour2016neural} and \citet{dong2017attention}. In particular, the 4000 most frequent words are preserved, with all other words treated as unknowns. The assessment scores will be converted back to their original range during evaluation. We use Quadratic Weighted Kappa (QWK) to evaluate our model. QWK is not only the official criteria of ASAP corpus, but also adopted as evaluation metric in \citet{rahimi2014automatic,taghipour2016neural,dong2017attention,rahimi2017assessing,zhang2017word} for both ASAP and RTA corpora. 

We use 5-fold cross-validation because both RTA and ASAP corpora have no released labeled test data. We split all corpora into 5 folds. For the ASAP corpus, the partition is the same as the setting presented by \citet{taghipour2016neural}. For the RTA corpus, since there is no prior work to split the corpus, we separate it into 5 folds randomly. In each fold, 60\% of the data are used for training, 20\% of the data are the development set, and 20\% of the data are used for testing.

To select the best model, we trained each model on 100 epochs and evaluated on the development set after each epoch. The best model is the model with the best QWK on the development set. This is done five times, once for each partition in the cross-validation. Then the average QWK score from these five evaluations on the test set is reported. Paired t-tests are used for significance tests with $p<0.05$. Table~\ref{hyper} shows all hyper-parameters for training.

The code of SELF-ATTN are provided by \citet{dong2017attention}, they used Keras \citep{chollet2015keras} 1.1.1 and Theano \citep{2016arXiv160502688short} 0.8.2 as the backend. Because we are using Keras 2.1.3 and TensorFlow \citep{tensorflow2015-whitepaper} 1.4.0 as the backend, we ran all experiments with our frameworks. Therefore, the numbers of SELF-ATTN have small differences to the numbers reported by the baseline model.

For non-neural network baselines, we introduce the SVR and BLRR baselines presented by \citet{phandi2015flexible} for the ASAP corpus, and SG baseline presented by \citet{zhang2017word} for the RTA corpus. 

SVR and BLRR models use Enhanced AI Scoring Engine (EASE)\footnote{https://github.com/edx/ease} to extract four types of features, such as length, part of speech, prompt, and the bag of words. Then they use SVR and BLRR as the classifiers, respectively. We do not perform any significance test on both SVR and BLRR because we do not have detailed experiment data. Therefore, we only report the result presented in \citet{phandi2015flexible}.

SG model extracts evidence features based on hand-crafted topic and example lists, and uses random forest tree as the classifier. We follow the same data partition. However, we only use the training set for training and the testing set for testing while ignoring the development set so that we can perform the same paired t-tests in the experiments.

\begin{table}[h]
\begin{center}
\begin{tabular}{|c|c|c|}
\hline \bf Layer & \bf Parameter Name & \bf Value  \\ \hline
Embedding & Embedding dimension & 50 \\ \hline
Word-CNN & Kernel size & 5 \\
& Number of filters & 100 \\ \hline
Sent-LSTM & Hidden units & 100 \\ \hline
Modeling & Hidden units & 100 \\ \hline
Dropout & Dropout rate & 0.5 \\ \hline
Others & Epochs & 100 \\
& Batch size & 100 \\
& Initial learning rate & 0.001 \\
& Momentum & 0.9 \\
\hline
\end{tabular}
\end{center}
\caption{\label{hyper} Hyper-parameters of training.}
\end{table}

\begin{table*}[h]
\begin{center}
\begin{tabular}{|c|c|c|c|c|c|}
\hline \bf Prompts & \bf SVR & \bf BLRR & \bf SG & \bf SELF-ATTN & \bf CO-ATTN  \\ \hline
$RTA_{MVP}$ & NA & NA & 0.653 & 0.681$\dagger$ & \bf 0.697$\ast\dagger$ \\
$RTA_{Space}$ & NA & NA & 0.632 & 0.669$\dagger$ & \bf 0.684$\ast\dagger$ \\
$ASAP_3$ & 0.630 & 0.621 & NA & 0.677 & \bf 0.697$\ast$ \\
$ASAP_4$ & 0.749 & 0.784 & NA & 0.807 & \bf 0.809 \\
$ASAP_5$ & 0.782 & 0.784 & NA & 0.806 & \bf 0.815 \\
$ASAP_6$ & 0.771 & 0.775 & NA & 0.809 & \bf 0.812 \\
\hline
\end{tabular}
\end{center}
\caption{\label{results} The performance (QWK) of the baselines and our model. $\ast$ indicates that the model QWK is significantly better than the SELF-ATTN ($p<0.05$). $\dagger$ indicates that the model QWK is significantly better than the SG ($p<0.05$). The best results in each row are in bold.}
\end{table*}

\section{Results}
We first examine H1. The results shown in Table~\ref{results} support this hypothesis. The CO-ATTN model yields higher performance than the SELF-ATTN model on all ASAP prompts. 
However, the CO-ATTN model only significantly outperforms the SELF-ATTN model on Prompt 3.

Second, we examine H2. Again, the results shown in Table~\ref{results} support this hypothesis. The CO-ATTN model yields higher performance than the SELF-ATTN model, significantly on both of the RTA corpora. 

Last, we examine H3. The results shown in Table~\ref{results} still support this hypothesis. The CO-ATTN model yields higher performance than all non-neural network baselines.

The results show that in our tasks, the neural network approaches are better than non-neural network baselines. One possible reason is the final representation of the essay from neural network contains more information. However, some of the information might be ignored by hand-crafted features. For example, the importance of different evidence in RTA task is not considered in the SG model. It treats all evidence equally. However, the neural network models capture this information automatically. 

Apparently, the CO-ATTN model performs better in the RTA tasks, because it always significantly outperforms the SELF-ATTN model. One possible reason is that the RTA task only considers the Evidence score. The CO-ATTN model is more suitable for the Evidence score prediction task because it can find pieces of evidence that appear in both students' essays and the source article better. In contrast, the SELF-ATTN model only considers students' essays associated with the scores. In this case, if a piece of evidence is not mentioned by students, this data-driven model cannot distinguish it. Consequently, some important pieces of evidence will be assigned to a lower weight. However, the CO-ATTN model considers not only the students' essays but also the source article. In other words, if an important piece of evidence is not mentioned by too many students, but it is in the source article, the CO-ATTN model will assign this sentence higher attention. 

In the ASAP holistic score prediction task, although we still see a benefit in using the CO-ATTN model, it is reduced. In this case, the benefit we saw in the Evidence dimension from the CO-ATTN model becomes less significant because the model also needs to consider more aspects of the essay, such as organization, grammar mistakes, and so on. Our results suggest that  the co-attention mechanism of the CO-ATTN model cannot capture  these aspects significantly better than the SELF-ATTN model. Therefore, the CO-ATTN model only significantly outperforms the SELF-ATTN model on Prompt 3.



\section{Discussion}

In Table~\ref{sents}, we list 10 sentences from student $RTA_{MVP}$ essays and their associated  attention scores. Because we have a list of examples manually extracted by our experts as important evidence from the $RTA_{MVP}$ source article, examining RTA data helps us understand the attention score assigned by our model. Bolded are examples extracted by the expert from the source article that the student includes in the essay. A lower attention score means this sentence is less important. Otherwise, the score is high. As we can see, sentences 1, 2, 3, and 4 are low attention sentences, sentences 5, 6, and 7 are mid attention sentences, and sentences 8, 9, and 10 are high attention sentences. The attention scores reflect the importance of these sentences accurately. 

Sentence 1 is a short and general sentence related to the source article, but it has no specific evidence from it. Sentence 2 even has no content related to the source article. Sentence 3 has many details related to the source article. However, it still has no evidence directly from the source article. Sentence 4 mentions ``\textit{The author did convince me that winning the fight against poverty is achievable in our lifetime}'' which comes from both the prompt and the source article, but this statement is so general that almost every student mentions this statement in the essay which makes this statement not distinguishable. For these reasons, these four sentences receive low attention scores.

Although sentence 5 is short, it mentions one piece of evidence. Sentence 6 talks about farming which is a topic from the source article. 
In the article, the things listed in this sentence are things the farmer needs to worry about. However, this sentence indicates ``\textit{the farmer don't have to worry}'' because of the MVP project. Sentence 7 also mentions conditions of hospitals nowadays. However, it mentions not only water but also electricity which is more than Sentence 5. For these reasons, these three sentences receive mid attention scores from low to high.

The last three sentences receive high attention scores because they all use more pieces of evidence directly from the source article. Sentence 8 talks about the school, and Sentence 9 talks about the hospital. Sentence 10 talks about farming. However, sentence 10 receives the highest attention score, because it mentions evidence from both before and after the MVP project. 

\begin{table}[h]
\begin{center}
\small
\begin{tabular}{|p{0.05\linewidth}|p{0.60\linewidth}|p{0.15\linewidth}|}
\hline \bf No. & \bf Sentences & \bf Attention  \\ \hline
1 & Life in Kenya is hard. & 0.00173 \\ \hline
2 & In this essay I will give my top 3 reasons why. & 0.00174 \\ \hline
3 & Because like I said, we have more advanced \& better \& more qualified materials than them, and these days kids \& adults are spoiled, we have phones stores, houses \& even shoes and clothes. & 0.00243 \\ \hline
4 & The author did {\bf convince me that winning the fight against poverty is achievable in our lifetime} because she showed me how many people in Sauri, Kenya need our help against poverty. & 0.00229 \\ \hline 
5 & {\bf Water is connected to the hospitals.} & 0.02936 \\ \hline
6 & So the farmer don't have to worry all the time that him or his {\bf family won't have enough food} to eat and the farmer have to worry that their kids will get hungry and then sick. & 0.05580 \\ \hline
7 & {\bf The hospital aslo has water and electricity}. & 0.07746 \\ \hline
8 & Also, there were {\bf no school fees}, and {\bf the school now serves lunch} for the students because they {\bf didn't have any midday meals} to provide them with {\bf energy they need} to help them with the rest of their days. & 0.19483 \\ \hline
9 & In 2008 though, when they {\bf checked for progress}, {\bf the hospital had medicine}, {\bf free of charge}, with {\bf running water and electricty}. & 0.20177 \\ \hline
10 & Also {\bf farmers could not afford fertilizer and irrigation} but now {\bf they placed irrigation} and have them {\bf fertilizer for the crops}. & 0.25855 \\
\hline
\end{tabular}
\end{center}
\caption{\label{sents} Example attention scores of essay sentences.}
\end{table}

From these sentences, we can also see that the attention score depends on neither the length of the sentence nor only the specificity of the sentence. It instead depends on how many important pieces of evidence  there are in the sentence. For example, Sentence 3 is long and talks about some details of our modern life. Although it also talks about quality materials or better housing and clothing compared to people living in Kenya, it receives a low attention score because there is no specific evidence directly from the source article. 
In contrast, Sentence 9 is shorter than Sentence 3. However, it receives a higher attention score because it mentions many pieces of evidence from the source article.


Overall, the CO-ATTN model seems to capture the importance of sentences by assigning reasonable attention scores based on the relevance of the sentence to the source article.

\section{Conclusion and Future Work}

In this paper, we presented a co-attention based neural network model that outperforms a  state of the art attention based neural network model for essay scoring, not only for  RTA Evidence assessment but also for holistic assessment of ASAP source-dependent responses. Advantages of our model are that it does not need any expert preprocessing of the source article; the input of this model is only the raw student essay and its source article. Moreover, our model somewhat captures the importance of different pieces of evidence, although it is not specifically designed for this purpose. However, quantitative experiments that can answer whether the attention scores are correlated to the importance of different pieces of evidence need to be done. Also, this leads to an interesting future investigation, development of a neural network approach that both has an acceptable score prediction, and can simultaneously generate evidence lists from the source article. Another interesting future investigation could be examining the ability of this model to generalize to a new prompt. 


\section*{Acknowledgments}
We would like to show our appreciation to Tazin Afrin and Luca Lugini for their comments on an earlier version of the paper. We are also immensely grateful to every member of the RTA group for sharing their pearls of wisdom with us.

The research reported here was supported, in whole or in part, by the Institute of Education Sciences, U.S. Department of Education, through Grant R305A160245 to the University of Pittsburgh. The opinions expressed are those of the authors and do not represent the views of the Institute or the U.S. Department of Education.

\bibliography{naaclhlt2018}
\bibliographystyle{acl_natbib}

\end{document}